\pdfoutput=1

\documentclass[11pt]{article}

\usepackage{acl}

\usepackage{times}
\usepackage{latexsym}

\usepackage[T1]{fontenc}

\usepackage[utf8]{inputenc}

\usepackage{microtype}

\usepackage{xcolor}

\usepackage{cleveref}
\crefname{section}{\S}{\S\S}
\Crefname{section}{\S}{\S\S}
\crefname{sec}{\S}{\S\S}
\Crefname{sec}{\S}{\S\S}
\crefname{table}{Table}{}
\crefname{figure}{Fig.}{Figs.}
\crefname{algorithm}{Algorithm}{}
\crefname{algorithm}{Algorithm}{}
\crefname{line}{Line}{}
\crefname{appendix}{App.}{}
\crefformat{section}{\S#2#1#3}
\crefname{thm}{Theorem}{}
\crefname{cor}{Corollary}{}
\crefname{prop}{Proposition}{}
\crefname{def}{Definition}{}

\usepackage{xurl}
\usepackage{graphicx}
\usepackage{arydshln}
\usepackage{subfigure}
\usepackage{booktabs}
\title{\textsc{NaijaHate}: Evaluating Hate Speech Detection \\ on Nigerian Twitter Using Representative Data}

\author{Manuel Tonneau \textsuperscript{\rm 1, \rm 2, \rm 3}, 
        Pedro Vitor Quinta de Castro \textsuperscript{\rm 1, \rm 4},
        Karim Lasri \textsuperscript{\rm 1, \rm 5},
        \\
        {\bf Ibrahim Farouq} \textsuperscript{\rm 1, \rm 6},
        {\bf Lakshminarayanan Subramanian} \textsuperscript{\rm 3},\\
        {\bf Victor Orozco-Olvera} \textsuperscript{\rm 1},
        {\bf Samuel P. Fraiberger} \textsuperscript{\rm 1, \rm 3, \rm 7},\\
        \textsuperscript{\rm 1} The World Bank,
        \textsuperscript{\rm 2} University of Oxford,
        \textsuperscript{\rm 3} New York University \\
        \textsuperscript{\rm 4} Universidade Federal de Goiás
        \textsuperscript{\rm 5} Ecole Normale Supérieure \\
        \textsuperscript{\rm 6} Universiti Sultan Zainal Abidin 
        \textsuperscript{\rm 7} Massachusetts Institute of Technology
        }
\begin{document}
\maketitle
\begin{abstract}
To address the global issue of online hate, hate speech detection (HSD) systems are typically developed on datasets from the United States, thereby failing to generalize to English dialects from the Majority World. Furthermore, HSD models are often evaluated on non-representative samples, raising concerns about overestimating model performance in real-world settings. In this work, we introduce \textsc{NaijaHate}, the first dataset annotated for HSD which contains a representative sample of Nigerian tweets. We demonstrate that HSD evaluated on biased datasets traditionally used in the literature consistently overestimates real-world performance by at least two-fold. We then propose \textsc{NaijaXLM-T}, a pretrained model tailored to the Nigerian Twitter context, and establish the key role played by domain-adaptive pretraining and finetuning in maximizing HSD performance. Finally, owing to the modest performance of HSD systems in real-world conditions, we find that content moderators would need to review about ten thousand Nigerian tweets flagged as hateful daily to moderate 60\% of all hateful content, highlighting the challenges of moderating hate speech at scale as social media usage continues to grow globally. Taken together, these results pave the way towards robust HSD systems and a better protection of social media users from hateful content in low-resource settings.
\end{abstract}

\noindent \textcolor{red}{\textbf{Content warning:} This article contains illustrative examples of hateful content.}

\section{Introduction}

Social media came with the promise to connect people, increase social cohesion, and let everyone have an equal say. Yet, this technology has also been associated with online harms and detrimental effects on democracy \cite{lorenz2023systematic}. In particular, users are regularly exposed to hate speech\footnote{Hate speech is defined as any communication that disparages a person or a group based on the perceived belonging of a protected characteristic, such as race \cite{basile-etal-2019-semeval}.} \cite{vidgen2019much}, fueling fears of its impact on social unrests and hate crimes \citep{muller2021fanning}. While regulatory frameworks have compelled social media platforms to take action \citep{gagliardone2016mechachal}, hate speech detection (HSD) and moderation efforts have largely focused on the North American and European markets, prompting questions on how to efficiently tackle this issue in the Majority World \cite{milmo2021haugen, tonneau2024languages}. Our study focuses on Nigerian Twitter, a low-resource context which provides an opportunity to study online hate speech at the highest level \cite{ezeibe2021hate}. Exemplifying the issue, Twitter was banned by the Nigerian government between June 2021 and January 2022, following the platform’s deletion of a tweet by the Nigerian President in which he made provocative remarks about Biafran separatists \cite{maclean2021ban}. 
 
 The challenges in developing HSD systems are two-fold. First, hateful content is infrequent -- approximately $0.5\%$ of posts on US Twitter \cite{jimenez2021economics} -- creating an obstacle to generating representative annotated datasets at a reasonable cost. 
 To alleviate this issue, models are developed on curated datasets by oversampling hateful content matching predefined keywords \cite{davidsonetal2017}, or by using active learning to maximize performance for a given annotation cost \cite{kirk-etal-2022-data, markov2023holistic}. These sampling choices generate biases in evaluation datasets \cite{wiegand-etal-2019-detection,nejadgholi-kiritchenko-2020-cross}, raising questions on the generalizability of HSD models to real-world settings. 

Second, while a plethora of HSD modeling options are available, it is unclear how well they adapt to a new context. Although few-shot learners are appealing for requiring no or few finetuning data, the evidence on their performances relative to supervised HSD baselines is mixed \cite{plazadelarco2023leveraging, guo2024investigation}. Supervised learners are typically finetuned on US data \cite{tonneau2024languages} and tend to not generalize well to English dialects spoken in the Majority World \cite{ghosh-etal-2021-detecting}. Finally, while further pretraining existing architectures to adapt them to a new context is known to increase performance on downstream tasks \cite{gururangan-etal-2020-dont}, it is unclear whether highly specific contexts require a custom domain adaptation. Overall, questions remain on the extent to which available HSD methods perform when adapted to a low-resource context \cite{li2021achieving}. 

In this work, we present \textsc{NaijaHate}, a dataset of 35,976 Nigerian tweets annotated for HSD, which includes a representative evaluation sample to shed light on the best approach to accurately detect hateful content in real-world settings. We also introduce \textsc{NaijaXLM-T}, a pretrained language model adapted to the Nigerian Twitter domain. We demonstrate that evaluating HSD models on biased datasets traditionally used in the literature largely overestimates performance on representative data (83-90\% versus 34\% in average precision). We further establish that domain-adaptive pretraining and finetuning leads to large HSD performance gains on representative evaluation data over both US and Nigerian-centric baselines. We also find that finetuning on linguistically diverse hateful content sampled through active learning significantly improves performance in real-world conditions relative to a stratified sampling approach.
Finally, we discuss the cost-recall tradeoff in moderation and show that having humans review the top 1\% of tweets flagged as hateful allows to moderate up to 60\% of hateful content on Nigerian Twitter, highlighting the limits of a human-in-the-loop approach to content moderation as social media usage continues to grow globally. 

Therefore, our main contributions are \footnote{The dataset and the related models can be found at \url{https://github.com/worldbank/NaijaHate}.}: 
\vspace{-2mm}
\begin{itemize}
\setlength\itemsep{0em}
\item \textsc{NaijaHate}, a dataset which includes the first representative evaluation sample annotated for HSD on Nigerian Twitter;
\item \textsc{NaijaXLM-T}, a pretrained language model adapted to the Nigerian Twitter domain;
\item an evaluation on representative data of the role played by domain adaptation and training data diversity and a discussion of the feasibility of hateful content moderation at scale.
\end{itemize}

\section{Related work}

\subsection{Nigerian hate speech datasets}
While existing English hate speech datasets are primarily in US English \cite{tonneau2024languages}, mounting evidence highlights the limited generalizability of learned hate speech patterns from one dialect to another \cite{ghosh-etal-2021-detecting, castillo-lopez-etal-2023-analyzing}. In this context, recent work has developed hate speech datasets tailored to the Nigerian context; however, the latter either focused on one form of hate speech \cite{aliyu2022herdphobia}, specific languages \cite{adam2023detection, aliyubeyond}, or particular events \cite{ndabula2023detection, ilevbare2024ekohate}. To the best of our knowledge, our work is the first to construct a comprehensive dataset annotated for hate speech for the entire Nigerian Twitter ecosystem, covering both the diversity of languages and hate targets.

\subsection{Hate speech detection}

HSD methods fall into three categories: rule-based \cite{mondal2017measurement}, supervised learning \cite{davidsonetal2017}, and zero-shot or few-shot learning (ZSL) \cite{nozza-2021-exposing}. Rule-based methods rely on predefined linguistic patterns and therefore typically achieve very low recall. Additionally, supervised learning require annotated datasets which are usually scarce in Majority World contexts. While recent advancements in ZSL could potentially circumvent the need to produce finetuning data for supervised learning, existing evidence on the relative performance of the two approaches is mixed \cite{plazadelarco2023leveraging,guo2024investigation}. 

A major shortcoming of this literature is that HSD is typically evaluated on curated non-representative datasets whose characteristics greatly differ from real-world conditions \cite{wiegand-etal-2019-detection,nejadgholi-kiritchenko-2020-cross}, raising concerns about overestimating HSD performance \cite{arango2019hate}. While past work has annotated representative samples for HSD, the samples have either not been used solely for evaluation \cite{vidgen-etal-2020-detecting} or originate from largely unmoderated platforms, such as Gab \cite{kennedy2022introducing}, where hatred is much more prevalent and therefore HSD is an easier task. To the best of our knowledge, we provide the first evaluation of HSD on a representative evaluation sample from a mainstream social media platform, Twitter, where hate content is rare.

\subsection{Hate speech moderation}

Most large social media platforms prohibit hate speech \cite{singhal2023sok} and have done so since their inception \cite{gillespie2018custodians}. 
However, with millions of daily social media posts,  enforcing this policy is a very challenging task. While such scale has motivated the use of algorithmic detection methods \cite{gillespie2020content}, these approaches have raised concerns related to the fairness and potential biases their use imply in moderation decisions \cite{gorwa2020algorithmic}. As a middle ground, recent work has proposed a human-in-the-loop approach \cite{lai2022human}, where a model flags content likely to infringe platform rules, which is then reviewed by humans who decide whether or not to moderate it. Albeit promising, it remains unclear whether this process is scalable both from a cost and a performance standpoint. To fill this gap, we provide the first estimation of the feasibility of a human-in-the-loop approach in the case of Nigerian Twitter. 

\vspace{3mm}

\section{Data}\label{sec:data}
\begin{figure}[h!]
    \centering
        \includegraphics[width=0.5\textwidth]{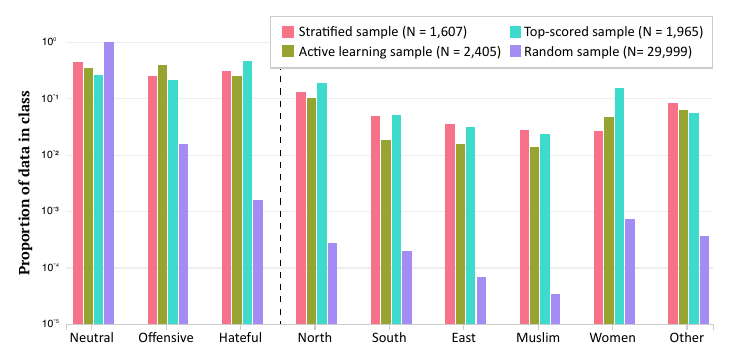}
\caption{Proportion of data in each class, showing the composition of the hateful class across hate target categories. }
\label{fig:descriptive_stats}
\end{figure}

\subsection{Data collection}
We use the Twitter API to collect a dataset containing 2.2 billion tweets posted between March 2007 and July 2023 and forming the timelines of 2.8 million users with a profile location in Nigeria.\footnote{The dataset contains 13.9 billion tokens and 525 million unique token, for a total of 89GB of uncompressed text.} We iteratively collect the timeline of users with a profile location in Nigeria being mentioned in the timeline of other Nigerian users until no additional Nigerian users were retrieved, ensuring maximum coverage of the Nigerian ecosystem. This Nigerian Twitter dataset is mostly constituted of English tweets (77\%) followed by tweets in Nigerian Pidgin -- an English-based creole widely spoken across Nigeria -- (7\%), tweets mixing English and Pidgin (1\%), tweets in Hausa (1\%) and tweets in Yoruba (1\%) (Table \ref{tab:language_distribution}). We then draw two distinct random samples of 100 million tweets each, one for model training and the other one for evaluation. 

\subsection{Annotation}

We recruit a team of four Nigerian annotators, two female and two male, each of them from one of the four most populated Nigerian ethnic groups -- Hausa, Yoruba, Igbo and Fulani. 

We follow a \textit{prescriptive} approach \cite{rottger-etal-2022-two} by instructing annotators to strictly adhere to extensive annotation guidelines describing our taxonomy of hate speech (detailed in \S \ref{app:annotation-guidelines}). Following prior work \cite{davidsonetal2017, mathew2021hatexplain}, HSD is operationalized by assigning tweets to one of three classes: (i) \textit{hateful}, if it contains an attack on an individual or a group based on the perceived possession of a protected characteristic (e.g., gender, race) \cite{basile-etal-2019-semeval}, (ii) \textit{offensive}, if it contains a personal attack or an insult that does not target an individual based on their identity \cite{zampieri-etal-2019-predicting}, or (iii) \textit{neutral} if it is neither hateful nor offensive. If a tweet is labeled as hateful, it is also annotated for the communities being targeted (Table \ref{tab:target_hate_examples}). 

In the next sections \S\ref{subsec:training_sample} and \S\ref{subsec:eval_samples}, we detail the different samples that were annotated. Each tweet in these samples was labeled by three annotators. For the three-class annotation task, the three annotators agreed on 90\% of all labeled tweets, two out of three agreed in 9.5\% of cases, and all three of them disagreed in 0.5\% of cases (Krippendorff's $\alpha$ = 0.7).

\begin{table}[hbt!]
    \centering
    \resizebox{0.5\textwidth}{!}{
    \begin{tabular}{lll}
        \hline
        \textbf{Label} & \textbf{Target} & \textbf{Examples} \\
        \hline

        & North & My hate for northern people keeps growing \\
        \cline{2-3}
        & South & You idiotic Southerners fighting your own \\
        \cline{2-3}
        & East &  IPOBs are animals....They lack tact or strategy. \\
        \cline{2-3}
        Hateful & Muslim & Muslim baboons and their terrorist religion. \\
        \cline{2-3}
         & Women & Nobody believes this \textit{ashawo} (prostitute) woman \\
        \hline
        Offensive & None & Stop spewing rubbish, \textit{mumu} (fool).\\ \hline
        Neutral & None & She don already made up her mind sha. \\ \hline 
    \end{tabular}}
    \caption{Examples of tweets for each class. Offensive tweets have no target as they do not target an identity group. Non-English words are displayed in italic with their English translation in parenthesis.}
    \label{tab:target_hate_examples}

\end{table}

\subsection{Training samples}\label{subsec:training_sample}

\paragraph{Stratified sample} Due to the rarity of hateful content, sampling tweets randomly would result in a highly imbalanced set. Indeed, the prevalence of hate speech on large social media platforms typically ranges from 0.003\% to 0.7\% depending on the platform and timeframe \cite{gagliardone2016mechachal, mondal2017measurement, jimenez2021economics}. To circumvent this issue, we follow previous work \cite{davidsonetal2017} by oversampling training examples containing keywords expected to be associated with hate. We handpick a list of 89 hate-related keywords combining hate speech lexicons and online dictionaries \cite{ferroggiaro2018social, farinde2020socio, udanor2019combating}. We also identify 45 keywords referring to communities frequently targeted by hate in the Nigerian context \cite{pate2020fake}, due to their ethnicity (Fulani, Hausa, Herdsmen\footnote{Herdsmen are not a ethnic group per se but this term refers exclusively to Fulani herdsmen in the Nigerian context, hence the categorization as an ethnic group.}, Igbo, Yoruba), religion (Christians, Muslims), region of origin (northerners, southerners, easterners), gender identity or sexual orientation  (women, LGBTQ+) \cite{onanuga2023arewaagainstlgbtq}. We then annotate 1,607 tweets from the training sample that are stratified by community-related and hate-related keywords (see \S\ref{app:stratified-sample}). Owing to the rarity of some individual targets and for descriptive purposes, we bundle targets into 6 target categories: North (Hausa, Fulani, Herdsmen, Northerner), South (Southerner, Yoruba), East (Biafra, Igbo, Easterner), Muslims, Women and Other (Christians and LGBTQ+). Stratified sampling indeed enables to reduce the imbalance in the training data (Figure \ref{fig:descriptive_stats}): the resulting shares of tweets labeled as neutral, offensive and hateful are respectively equal to 50\%, 17\%, and 33\%.

\paragraph{Active learning sample} While stratified sampling makes it possible to oversample hateful content in the training data, it is constrained by a predefined vocabulary which limits the coverage and diversity of the positive class. As an alternative, we employ certainty sampling \cite{attenberg2010unified} to annotate a second sample of training examples. Certainty sampling is an active learning method that focuses the learning process of a model on instances with a high confidence score of belonging to the minority class, spanning a more diverse spectrum of examples. We generate additional training instances in four steps: (i) we start by finetuning Conversational BERT \cite{burtsev-etal-2018-deeppavlov} on the stratified sample; (ii) we then deploy the finetuned model on the training sample of 100 million tweets; (iii) next, we label an additional 100 tweets sampled from the top 10\% tweets from the training sample in terms of confidence score; and finally, (iv) we incorporate the additional labels into Conversational BERT's finetuning sample. We repeat this process 25 times, thereby producing a set of 2,500 training examples, of which 2,405 have a majority label after annotation. We find that certainty sampling produces about the same proportion of observations from the hateful class (25\% versus 31\%) as stratified sampling (Figure \ref{fig:descriptive_stats}). However, it enables to generate more diversity in the hateful class: the proportion of training examples that do not contain any seed keywords\footnote{i.e., keywords used for stratified sampling}, the proportion of unique tokens and the average pairwise GTE embedding distance \cite{li2023towards} are consistently larger in the active learning sample compared to the stratified sample (Table \ref{tab:diversity}).


\begin{table}[htbp]
\centering
\resizebox{0.5\textwidth}{!}{%
\begin{tabular}{@{}lcccc@{}}
\toprule
 &
  \textbf{Stratified} &
  \textbf{Active learning} &
  \textbf{Top-scored} &
  \textbf{Random} \\ 
\midrule
\begin{tabular}[c]{@{}l@{}}Proportion of \\ tweets not\\ containing\\ seed keywords (\%)\end{tabular} & 7.5 $\pm$ 1.2 & 72.5 $\pm$ 1.8 & 70.8 $\pm$ 1.5 & 93.8 $\pm$ 3.5 \\ 
\midrule
\begin{tabular}[c]{@{}l@{}}Proportion of \\ unique tokens (\%)\end{tabular} & 32.2 $\pm$ 0.4 & 33.3 $\pm$ 0.4 & 29 $\pm$ 0.3 & 61.5 $\pm$ 1.6 \\ 
\midrule
\begin{tabular}[c]{@{}l@{}}Average pairwise \\ embedding \\ distance\end{tabular} & 0.139 $\pm$ 0.0 & 0.152 $\pm$ 0.0 & 0.159 $\pm$ 0.0 & 0.172 $\pm$ 0.001 \\ 
\bottomrule
\end{tabular}%
}
\caption{Diversity metrics for the hateful class across datasets. Active learning enables to generate more diversity in the training data, bringing them closer to the representative random sample. Metrics are reported with standard errors.}
\label{tab:diversity}
\end{table}

\subsection{Evaluation samples}\label{subsec:eval_samples}

\paragraph{Top-scored sample} To evaluate models' performance in real-world conditions, we start by testing how they behave in the presence of a distribution shift. We first train each supervised model considered in this study on the union of the stratified and the active learning sample, deploy it on the random sample of 100 million tweets used for evaluation and annotate 200 high-scored tweets. We repeat this process for the 10 models evaluated in this study (see \S \ref{sec:experimental_setup} for more details) and combine all the high-scored tweets, yielding a pooled sample of 2,000 tweets, of which 1,965 have a majority label after annotation. The shares of tweets labeled as neutral, offensive and hateful are respectively equal to 28\%, 22\% and 50\% (Figure \ref{fig:descriptive_stats}). This approach traditionally used in information retrieval enables to evaluate the performance of each model on a large dataset containing a high and diverse proportion of positive examples discovered by qualitatively different models, and whose distribution differs from that of the training data \cite{voorhees2005trec}. 

\vspace{2mm}

\paragraph{Random sample} Finally, we annotate a random sample of 30,000 tweets, of which 29,999 have a majority label after annotation, to evaluate HSD models on a representative dataset of Nigerian tweets. As expected, we discover that the prevalence of hateful content is very low: approximately $0.16\%$ and $1.6\%$ of tweets are labeled as hateful and offensive, respectively (Figure \ref{fig:descriptive_stats}). In addition, we find that the diversity within the positive class in the random sample is larger than in the training samples (Table \ref{tab:diversity}).

\section{Experimental setup}\label{sec:experimental_setup}
In this study, our experiments aim to determine the impact of domain adaptation -- both for pretraining and finetuning -- on real-world performance. For this purpose, we evaluate the performance of supervised learning with language models pretrained and finetuned in various domains that we present below. Additionally, we test recent off-the-shelf general-purpose models in a zero-shot setting, specifically \textit{GPT-3.5}\footnote{\url{https://openai.com/blog/chatgpt}}. We finally benchmark the finetuned models against Perspective API \cite{lees2022new}, a widely-deployed toxic language detection system relying on BERT-based supervised learning and for which the finetuning data is not public. 

\vspace{2mm}

\paragraph{Finetuning domain}
We experiment with four finetuning datasets: \textsc{HateXplain} \cite{mathew2021hatexplain}, which contains US English posts from Twitter and Gab annotated for HSD; \textsc{HERDPhobia} \cite{aliyu2022herdphobia}, a dataset of Nigerian tweets annotated for hate against Fulani herdsmen; \textsc{HSCodeMix} \cite{ndabula2023detection}, containing Nigerian tweets posted during the EndSARS movement and the 2023 Nigerian presidential election and annotated for general hate speech; and finally \textsc{NaijaHate}, our dataset presented in \S \ref{sec:data}. 

\vspace{2mm}

\paragraph{Pretraining domain}

We introduce \textsc{NaijaXLM-T} (full), an XLM-R model \cite{conneau-etal-2020-unsupervised} further pretrained on 2.2 billion Nigerian tweets for one epoch. We compare its performance relative to BERT-based models pretrained in three different domains:

\begin{itemize}
\setlength\itemsep{0em}
    \item the general domain, which includes a variety of sources such as books and news, both in English (DeBERTaV3 \cite{he2021debertav3}) and in multilingual settings (XLM-R \cite{conneau-etal-2020-unsupervised}, mDeBERTaV3 \cite{he2021debertav3})
    \item the social media domain, both in English (Conversational BERT \cite{burtsev-etal-2018-deeppavlov}, BERTweet \cite{nguyen-etal-2020-bertweet}) and in multilingual settings (XLM-T \cite{barbieri-etal-2022-xlm}) 
    \item social media posts marked as offensive, abusive or hateful (HateBERT \cite{caselli-etal-2021-hatebert} and fBERT \cite{sarkar-etal-2021-fbert-neural})
    \item the African domain (AfriBERTa \cite{ogueji-etal-2021-small}, Afro-XLM-R \cite{alabi-etal-2022-adapting} and XLM-R Naija \cite{adelani-etal-2021-masakhaner}). 
\end{itemize}

\noindent Differences in performance across models may be explained by several factors including not only the pretraining domain, but also pretraining data size and preprocessing, model architecture and hyperparameter selection. While it is hard to account for the latter as they are rarely made public, we estimate the impact of the pretraining domain on performance, holding pretraining data size and model architecture constant. To do so, we introduce \textsc{NaijaXLM-T (198M)}, an XLM-R model further pretrained on a random sample of 198 million Nigerian tweets, matching the amount of data used to pretrain XLM-T on multilingual tweets. We adopt the same preprocessing as for XLM-T by removing URLs, tweets with less than three tokens, and running the pretraining for one epoch.

\vspace{2mm}

\paragraph{Evaluation} HSD models are evaluated by their average precision for the hateful class, a standard performance metric in information retrieval which is particularly well-suited when class imbalance is high. For supervised learning, we perform a 90-10 train-test split and conduct a 5-fold cross-validation with five learning rates ranging from 1e-5 to 5e-5 and three different seeds. The train-test split is repeated for 10 different seeds, and the evaluation metrics are averaged across the 10 seeds.

\begin{table*}[h]
\centering
\begin{tabular}{|c|c|c|c|c|c|}
\hline
Pretraining data & Finetuning data & Model & Holdout & Top-scored & Random \\
\hline
Multiple & None & GPT-3.5, ZSL & - & 60.3$\pm$2.7 & 3.1$\pm$1.2\\
domains & Mixed$^{*}$ & Perspective API & - & 60.2$\pm$3.5 & 4.3$\pm$2.6\\ \hline 
Social & \textsc{HateXPlain} & XLM-T & \textit{84.2 $\pm$ 0.6 }& 51.8 $\pm$ 0.7 & 0.6 $\pm$ 0.1\\
Media & \textsc{HERDPhobia}$^{*}$ & XLM-T & \textit{62.0 $\pm$ 2.3} & 68.9 $\pm$ 0.8 & 3.3 $\pm$ 0.6\\
& \textsc{HSCodeMix}$^{*}$ & XLM-T & \textit{70.5 $\pm$ 3.7} &  63.7 $\pm$ 1.1 & 1.9 $\pm$ 0.5 \\ \hline
Multiple & & DeBERTaV3 & \textbf{82.3 $\pm$ 2.3} & 85.3 $\pm$ 0.8 & \textbf{29.7 $\pm$ 4.1} \\
domains & & XLM-R & 76.7 $\pm$ 2.5 & 83.6 $\pm$ 0.8 & 22.1 $\pm$ 3.7 \\
& & mDeBERTaV3 & 29.2 $\pm$ 2.0 & 49.6 $\pm$ 1.0 & 0.2 $\pm$ 0.0 \\ \cline{0-0} \cline{3-6}
Social & \textsc{NaijaHate} & Conv. BERT & 79.2 $\pm$ 2.3 & 86.2 $\pm$ 0.8 & 22.6 $\pm$ 3.6 \\
media &  & BERTweet & \textbf{83.6 $\pm$ 2.0} & \textbf{88.5 $\pm$ 0.6} & \textbf{34.0 $\pm$ 4.4} \\
& Stratified  & XLM-T & 79.0 $\pm$ 2.4 & 84.5 $\pm$ 0.9 & 22.5 $\pm$ 3.7 \\ \cline{0-0} \cline{3-6}
Hateful social & + & HateBERT & 67.8 $\pm$ 2.9 & 78.6 $\pm$ 0.9 & 7.6 $\pm$ 2.1 \\
media & certainty & fBERT & 67.8 $\pm$ 2.8 & 78.9 $\pm$ 0.9 & 10.1 $\pm$ 2.6 \\ \cline{0-0} \cline{3-6}
African & sampling & AfriBERTa & 70.1 $\pm$ 2.7 & 80.1 $\pm$ 0.9 & 12.5 $\pm$ 2.8 \\
languages & (N=4012) & AfroXLM-R & 79.7 $\pm$ 2.3 & 86.1 $\pm$ 0.8 & 24.7 $\pm$ 4.0 \\
&  & XLM-R Naija & 77.0 $\pm$ 2.5 & 83.5 $\pm$ 0.8 & 19.1 $\pm$ 3.4\\ \cline{0-0} \cline{3-6}
Nigerian Twitter & & \textsc{NaijaXLM-T} (198M) & \textbf{83.0 $\pm$ 2.2} & \textbf{90.2 $\pm$ 0.6} & \textbf{33.1 $\pm$ 4.3} \\
 & & \textsc{NaijaXLM-T} (full) & \textbf{83.4 $\pm$ 2.1} & \textbf{89.3 $\pm$ 0.7} & \textbf{33.7 $\pm$ 4.5} \\
\hline
\end{tabular}
\caption{Average precision (in \%) for the hateful class across models and evaluation sets. Metrics are reported with 95\% bootstrapped confidence intervals. All supervised learning classifiers are framed as three-class classifiers, except the models trained on finetuning data marked with an asterisk as the latter is binary (hateful or not). Hyphens indicate the absence of a holdout set. Metrics in italic are calculated on holdout sets that are different from one another and from the \textsc{NaijaHate} holdout set.}
\label{tab:hate_speech_evaluation_v25}
\end{table*}


\section{Results}
\label{sec:results}

\subsection{Hate speech detection}

\paragraph{Evaluating on representative data} 

 We evaluate HSD performance on three datasets: the holdout sample from the train-test splits, the top-scored sample and the random sample described in \S \ref{subsec:eval_samples} (Table \ref{tab:hate_speech_evaluation_v25}). Overall, we observe that the ordering of models' performance remains stable across evaluation sets. However, the striking result is that across the wide range of models considered in this study, the average precision on the random sample (20.92\% on average) is substantially lower than that on the holdout (73.75\%) and top-scored sets (81.88\%). This finding highlights the risk of considerably overestimating classification performance when evaluating HSD models on a dataset whose characteristics greatly differ from real-world conditions. We now present more details on the impact of the learning framework, the pretraining domain and the finetuning domain on performance.

\paragraph{Learning framework} We find that in-domain supervised learning on the \textsc{NaijaHate} dataset (90.2\% average precision on the top-scored set and 34\% on the random set) generally largely outperforms GPT3.5-based zero-shot learning (ZSL) (60.3\% and 3.1\%) and out-of-domain supervised learning on existing US and Nigerian-centric hate speech datasets (68.9\% and 4.3\%). 
We also note that ZSL and out-of-domain supervised learning, such as Perspective API, perform on par with each other.  Given that the prompt used does not provide a definition of hate speech (App. \ref{app:off_the_shelf_models}), it implies that GPT3.5 has incorporated enough knowledge from pretraining and reinforcement learning with human feedback to conceptualize and categorize hate speech as well as models finetuned on thousands of examples. Still, it exhibits a rather low performance which is likely due to the predominance of US English in the pretraining data, making it hard to generalize to Nigerian English.

\paragraph{Pretraining domain} We find that the language and register models are pretrained on have a large impact on downstream performance.
In-domain pretraining on Nigerian Twitter, combining Nigerian dialects and informal social media language, generally outperforms the other models both on the top-scored and random samples, followed by models pretrained on social media and general purpose domains. This result also holds when keeping the architecture and pretraining data size constant, with NaijaXLM-T (198M) yielding significantly better performance than XLM-T. A notable exception to NaijaXLM-T's dominance is BERTweet, a RoBERTa model pretrained from scratch on English tweets, which is on par with NaijaXLM-T on all evaluation sets. Such performance may be explained by the predominance of English on Nigerian Twitter, granting an advantage to English-centric models such as BERTweet or DeBERTaV3. It is also plausible that BERTweet’s pretraining data contains some English tweets originating from Nigeria. Finally, BERTweet was pretrained from scratch on tweets, implying that its vocabulary is tailored to the social media lingo, contrary to the XLM-based models.

We also observe that pretraining in the social media domain generally yields larger improvements than in the African linguistic domain. For instance, XLM-R Naija, an XLM-R model further pretrained on news in Nigerian Pidgin English, produces a rather poor performance, especially on the random set (19.1\% average precision), which is likely due to differences between news and social media lingo as well as the limited share of tweets in Pidgin English. Still, pretraining on the social media domain does not ensure a high performance, as illustrated by the low performance of models pretrained on toxic social media posts, such as HateBERT and fBERT. This underperformance may be explained by the fact that these models are pretrained on US-centric posts that emanate partly from other platforms than Twitter (e.g., Reddit) and contain aggressive and offensive speech, which may differ from hate speech.

\paragraph{Finetuning domain} In-domain finetuning on the NaijaHate dataset outperforms out-of-domain finetuning on both US-centric (Perspective, HateXPlain) and Nigerian-centric (HERDPhobia and HSCodeMix) datasets. When inspecting classification errors, we find that XLM-T HateXPlain, which is finetuned on US data, flags as hateful tweets that contain words that are very hateful in the US but not necessarily in Nigeria. For instance, ``ya k*ke'' means ``How are you'' in Hausa while k*ke is an antisemitic slur in the US context. As a result, XLM-T HateXplain assigns very high hateful scores to tweets containing this sentence whereas XLM-T NaijaHate does not. While finetuning on Nigerian Twitter data yields better performance than on US data, it does not ensure high performance, as illustrated by the poor performance of XLM-T HERDPhobia and HSCodeMix. 
In the case of HERDPhobia, the focus on hate against Fulani herdsmen leads to a low performance on other types of hate such as misogyny, underlining the importance of designing a comprehensive annotation scheme covering the most prevalent types of hate.

\paragraph{Finetuning data diversity}

\begin{figure}[h!]
    \centering
        \includegraphics[width=7.75cm]{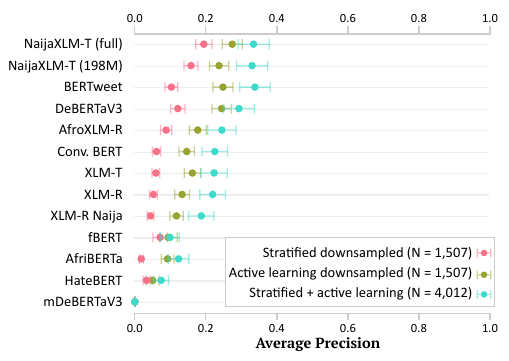}
\caption{Average precision on the random set across models trained on the downsampled stratified set, the downsampled active learning set and the full training set, composed of the stratified and active learning sets. Error bars indicate 95\% bootstrapped confidence intervals.}
\label{fig:diversity_random}
\end{figure}

In light of the higher diversity in the training data sampled through active learning compared to stratified sampling (Table \ref{tab:diversity}), we further investigate the role that finetuning data diversity plays on downstream performance. Specifically, we produce downsampled versions of the stratified and the active learning sets keeping dataset size and class distribution constant.  We report the results on the random sample in Figure \ref{fig:diversity_random} and on the other evaluation sets in Figure \ref{fig:diversity_rest} in the Appendix.

We find that finetuning on more diverse data significantly and consistently improves the average precision across models. The performance gains from diversity are particularly large for models that are not pretrained in the African linguistic domain, such as BERTweet and DeBERTaV3. We also discover that NaijaXLM-T significantly outperforms BERTweet on the less diverse stratified set. This finding indicates that the performance gains from in-domain pretraining may be particularly large when the finetuning data is less diverse, presumably because the lower diversity in the finetuning data is counterbalanced by a higher diversity and domain alignment in the pretraining data, allowing for a better generalization in real-world settings. 

\paragraph{Precision-recall tradeoff} We find that no classification threshold allows to reach a precision and recall that are both above 0.5 (Figure \ref{fig:pr_curve_random} in the Appendix), with the best possible F1 value for NaijaXLM-T (full) at 0.395. This high likelihood of error renders automatic moderation undesirable and calls for alternative moderation approaches, such as the human-in-the-loop moderation we present in the next section.

\subsection{Human-in-the-loop moderation}

\begin{figure}[h!]
    \centering
        \includegraphics[width=7.75cm]{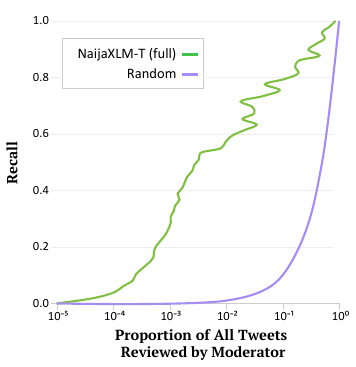}
\caption{Proportion of tweets flagged as hateful and reviewed by moderators versus recall. The random model flags tweets as hateful at random.}
\label{fig:moderation}
\end{figure}
In light of the low performance of hate speech classifiers on Nigerian real-world data, we explore the feasibility of a human-in-the-loop approach to hate speech moderation, in which content likely to contain hate is flagged by a classifier and then reviewed by humans. Instead of the traditional precision-recall tradeoff, human-in-the-loop moderation implies a \textit{cost-recall tradeoff}, where augmenting the recall comes at the cost of having more flagged content reviewed by humans. We represent this tradeoff by plotting the relationship between the share of all tweets flagged as hateful by a hate speech classifier, thus requiring human annotation in a human-in-the-loop moderation setting, and recall (Figure \ref{fig:moderation}). We provide more details in \S \ref{app:human-in-the-loop}.

We find that supervised learning enables to divide the amount of flagged tweets to be reviewed by a factor of 60 compared to a random baseline, with 1\% of the data to be sent for review in order to achieve a recall of 60\%. With an average daily flow of about 920,000 tweets on Nigerian Twitter in 2023, this translates to an average of 9,200 tweets to be reviewed daily, which is a feasible objective for a small team of moderators. However, increasing recall comes at a cost that quickly becomes prohibitively high, even for relatively small markets such as Nigeria: 10\% of posts would need to be reviewed daily in order to reach 76\% recall.

\section{Discussion}

\paragraph{Performance overestimation} We demonstrate that evaluating HSD on biased datasets leads to a large overestimation of real-world performance, the latter being rather low (34\% average precision). A possible reason for this gap is that hateful content is much more rare in real-world settings (0.16\% in our Nigerian Twitter case) compared to traditional evaluation sets (31\% in English hate speech datasets \cite{tonneau2024languages}), rendering its detection more challenging. This result expands on past work which pointed at the risk of overestimating HSD performance in real-world settings but did not quantify the latter \cite{arango2019hate,wiegand-etal-2019-detection,nejadgholi-kiritchenko-2020-cross}. 

\paragraph{Feasibility of moderation} Low real-world HSD performance has implications for hate speech moderation, making automated moderation unfeasible on top of being problematic for fairness and bias reasons \cite{gorwa2020algorithmic}. In this context, we investigate the feasibility of human-in-the-loop moderation, where content likely to be hateful is flagged by a model before being reviewed by humans. We highlight a cost-recall tradeoff, where a higher recall comes at the expense of increasing reviewing efforts and find that 60\% recall can be achieved by reviewing 1\% of all tweets. While reviewing 1\% of posts is a feasible goal for small online communities like Nigerian Twitter, it represents substantial costs for larger communities, both financial and in terms of reviewer harm \cite{steiger2021psychological}. We also find that increasing recall comes at a cost that quickly becomes prohibitively high for both small and large communities -- 10\% of posts need to be reviewed to reach a 76\% recall. This may partly explain the low hate speech removal rates on large social media platforms (e.g., 3-5\% on Facebook \cite{giansiracusa2021facebook}) and highlights the need for complementary approaches to support the moderation effort. 

\paragraph{Learning framework} In terms of HSD approaches, we find that in-domain supervised learning significantly outperforms both out-of-domain supervised learning and zero-shot learning. We also find that out-of-domain supervised learning and zero-shot learning perform on par with each other. These findings complement prior work underlining the superiority of supervised learning over zero-shot learning for HSD \cite{plazadelarco2023leveraging} by extending this result to a low-resource setting.

\paragraph{Pretraining domain} Further, the choice of pretraining model has a large impact on downstream performance. Pretraining on in-domain data that blends the noisy aspect of social media text with the linguistic domain of finetuning tasks yields significantly better performance than pretraining only on the former, even when holding pretraining data size and model architecture constant. Our results support the established finding that in-domain pretraining increases downstream task performance \cite{gururangan-etal-2020-dont} and complements it by underlining the importance of all relevant domains during pretraining, both in terms of register and linguistic focus. 

\paragraph{Finetuning domain} 

We find that in-domain finetuning outperforms out-of-domain finetuning. The main reason is that hateful statements are highly context-specific \cite{lee2023crehate, davani2024d3code}, leading to a limited generalizability of HSD across cultures, as pointed out in previous work \cite{ghosh-etal-2021-detecting}. It is therefore crucial to develop HSD datasets in low-resource contexts and our work contributes to this effort for Nigeria, which is largely under-represented in English HSD resources \cite{tonneau2024languages}.

\paragraph{Role of diversity} We observe that using diverse data acquired through active learning yields significant performance gains over stratified sampling.   This suggests that annotating a small stratified sample and acquiring a larger and more diverse dataset through active learning is preferable to only using stratified data. Our results align and complement past findings showing the benefits of active learning to maximize performance at a limited cost \cite{kirk-etal-2022-data}, including in extremely imbalanced settings like ours \cite{tonneau-etal-2022-multilingual}, and help better understand them through the prism of diversity. 

\section{Conclusion}

This work introduced \textsc{NaijaHate}, the largest HSD dataset to date in the Nigerian context and the first to contain a representative evaluation set. We also introduced \textsc{NaijaXLM-T}, the first pretrained language model tailored to Nigerian Twitter. Leveraging these resources, we provided the first evaluation of HSD in a real-world setting, finding that HSD performance is largely overestimated by traditional evaluation approaches. We finally presented the cost-recall tradeoff at play in human-in-the-loop moderation, where augmenting the recall comes with a higher annotation cost. We find that this moderation approach is effective at moderating a large share of all hateful content but that the associated annotation cost is prohibitive for large social media communities.

While the present work demonstrates the low real-world performance of HSD and the substantial cost of hate speech moderation, there are several possible directions to further improve this performance. Based on the hypothesis that hate is homophilous \cite{jiang2023social}, future work could use network features to improve HSD performance \cite{ahmed2022tackling}. Synthetic data could be used as well to further increase the number and diversity of examples to train models on \cite{khullar2024hate}. Future work could also evaluate performance at the target level to identify hate types that are harder to detect and develop resources for these. Further, the moderation analysis could be enhanced by taking popularity into account and measuring recall in terms of views of hateful content. Finally, future work could expand our study to other languages, countries and platforms in order to evaluate the external validity of our results and pinpoint areas where new resources are most needed.  

\vspace{2mm}

\section*{Limitations}

\paragraph{Dataset}

\textit{Low number of positives}:
The random set used to evaluate HSD in real-world settings contains 49 hateful examples, which is a very small number. This number is linked to the very low prevalence of hate speech in our dataset as well as our budget constraint which impeded us from further expanding the annotation effort. While statistically significant, we acknowledge that our performance results on the random set are necessarily more uncertain, as illustrated by the larger confidence intervals.

\textit{Limited generalizability to other platforms, timeframes and linguistic domains}:
The entirety of our dataset was sampled from one single social media platform for a long yet bounded timeframe. This limits the generalizability of models trained on our dataset to data from other social media platforms and covering other timespans. Our dataset is also specific to the Nigerian linguistic context and may exhibit poorer performance in other English dialects.

\textit{We do not exhaust all targets of hate}:
The selection of communities often targeted by hate speech and frequent on Nigerian Twitter necessarily leaves out of the analysis other communities even though they are targeted by online hate speech. In the annotation process, we observed for instance that South Africans, British people and Men are also targeted on Nigerian Twitter. 

\textit{Moderation prior to collection}: Our analysis of moderation considers that the hateful content in our random sample is representative of all hateful content on Nigerian Twitter. We acknowledge though that some hateful content may have been moderated by Twitter before we collected it and that our estimate of the prevalence of hate speech is necessarily a lower bound estimate. However, based on publicly available information, we suspect the enforcement of hate speech removal to be very low \cite{giansiracusa2021facebook}.

\paragraph{Experiments}

\textit{Other prompts could lead to different results}:
We craft a prompt using the terms ``hateful'' and ``offensive'' (see \cref{app:off_the_shelf_models} for details) which exhibit good performance in past research for HSD using ZSL \cite{plaza-del-arco-etal-2023-respectful}. We do not test other prompts and acknowledge that using other prompts may have an impact on classification performance.

\section*{Ethical considerations}

\paragraph{Annotator Wellbeing}

Annotators were provided with clear information regarding the nature of the annotation task before they began their work. They received a compensation of 12 US dollars per hour, which is above the Nigerian minimum wage.

\paragraph{Data Privacy}

We collected public tweets through the Twitter API according to its Terms and Services. To protect the identity of hateful users and their victims, we anonymize all tweets in our dataset upon release, replacing all user names by a fixed token @USER.

\section*{Acknowledgements}

We thank Daniel Barkoczi, Khashau Adenike Eleburuike, Niyati Malhotra, Haaya Naushan, Arinze Alex Okoyeugha, Luis Eduardo San Martin, Fatima Suleiman Sanusi and Philipp Zimmer for their excellent research assistance. We also thank Alice Grishchenko for outstanding visual design work. This work is part of a broader project conducted by the World Bank Development Impact (DIME) department. The study was supported by funding from the United Kingdom’s Foreign Commonwealth and  Development Office (FCDO) and the World Bank’s Research Support Budget. This work was also supported in part through the NYU IT High Performance Computing resources, services, and staff expertise. The findings, interpretations, and conclusions expressed in this paper are entirely those of the authors. They do not necessarily represent the views of the International Bank for Reconstruction and Development/World Bank and its affiliated organizations or those of the Executive Directors of the World Bank or the governments they represent.

\bibliography{anthology,custom}
\bibliographystyle{acl_natbib}

\clearpage

\appendix

\section{Experimental details}

\subsection{Data collection}\label{appendix:data_collection}

\subsubsection{Word lists}\label{appendix:word_lists}
In this section, we provide the detailed lists of slurs and communities specific to Nigerian hate speech (Table \ref{tab:list_hate_words}). Summary statistics on the number of words per category can be found in Table \ref{tab:summary_stats_word_lists}. 

\paragraph{Hate words}

We first build a list of 89 Nigeria-specific slurs, which are referred to as \textit{hate words} thereafter. To do so, we rely on lexicons from past work on the topic \cite{ferroggiaro2018social,udanor2019combating, farinde2020socio}, Nigerian online dictionaries such as Naija Lingo as well as local knowledge from our Nigerian colleagues. 
The final list contains two types of words: regular slurs (n=84) and words combining a slur and community name, such as ``fulanimal'' (n=5). The list of 84 regular slurs contains 28 Yoruba words, 26 English words, 12 Hausa words, 11 Pidgin words and 7 Igbo words. We detail the full list of hate words in Table \ref{tab:list_hate_words}.

\begin{table*}[h!]
\centering
\small
\scalebox{0.45}{
\begin{tabular}{|l|l|l|p{5cm}|p{5cm}|}
\hline
\textbf{Hate Keyword} & \textbf{Language} & \textbf{Translation} & \textbf{Source} \\
\hline
stupid & english & & \\
\hline
animal|animals & english & & \\
\hline
baboon|baboons & english & & \\
\hline
bastard|bastards & english & & \\
\hline
bitch|bitches & english & & \\
\hline
bum|bums & english & & \\
\hline
cockroach|cockroaches & english & & \\
\hline
coconut head|coconut heads & english & & \\
\hline
disgusting & english & & \\
\hline
dog|dogs & english & & \\
\hline
dumb|dumb & english & & \\
\hline
fanatic|fanatics & english & & \\
\hline
fool|fools & english & & \\
\hline
idiot|idiots & english & & \\
\hline
liar|liars & english & & \\
\hline
moron|morons & english & & \\
\hline
parasite|parasites & english & & \\
\hline
pig|pigs & english & & \\
\hline
primitive|primitives & english & & \\
\hline
rape|rapes|raping & english & & \\
\hline
scum|scums & english & & \\
\hline
shit|shits & english & & \\
\hline
slut|sluts & english & & \\
\hline
useless & english & & \\
\hline
vulture|vultures & english & & \\
\hline
whore|whores & english & & \\
\hline
aboki|abokai & hausa & "friend; used by a non-Hausa person may be derogatory" & \url{https://www.bellanaija.com/2020/04/twitter-aboki-derogatory-term/} \\
\hline
arne|arna & hausa & "pagan - used by muslims to reference christians in the north" & \\
\hline
ashana & hausa & prostitute & \url{http://naijalingo.com/words/ashana} \\
\hline
barawo|barayi & hausa & thief & \url{http://naijalingo.com/words/barawo} \\
\hline
bolo & yoruba & fool & \url{http://naijalingo.com/words/bolo} \\
\hline
kafir|kafirai & hausa & "used by muslims to refer to non-muslims" & \\
\hline
mallam|malamai & hausa & "teacher; used specifically in southern Nigeria in derogatory manner to refer to all Northerners; in Northern Nigeria, is used as a mark of respect" & \\
\hline
malo|malos & hausa & fool & \\
\hline
mugu & hausa & wicked/evil & \url{http://naijalingo.com/words/mugu} \\
\hline
mugun & hausa & fool & \url{http://naijalingo.com/words/mugun} \\
\hline
mungu & hausa & fool & \url{http://naijalingo.com/words/mungu} \\
\hline
wawa|wawaye & hausa & idiot & \\
\hline
zuwo & hausa & fool & \url{http://naijalingo.com/words/zuwo} \\
\hline
anuofia|ndi anofia & igbo & wild animal & \\
\hline
aturu & igbo & sheep & \citet{udanor2019combating} \\
\hline
efulefu|ndi fulefu & igbo & worthless man & \\
\hline
ewu & igbo & & \\
\hline
imi nkita & igbo & dog nose & \url{https://www.vanguardngr.com/2019/11/of-yariba-nyamiri-and-aboki/} \\
\hline
onye nzuzu|ndi nzuzu & igbo & & \\
\hline
onye oshi|ndi oshi & igbo & thief & \url{http://naijalingo.com/words/onye-oshi} \\
\hline
ashawo|ashawos & pidgin & prostitute & \url{http://naijalingo.com/words/ashawo} \\
\hline
ashewo|ashewos|awon ashewo & pidgin & prostitute & \url{http://naijalingo.com/words/ashewo} \\
\hline
ashy & pidgin & dirty & \\
\hline
mumu|mumus & pidgin & idiot & \\
\hline
mumuni & pidgin & very stupid person & \url{http://naijalingo.com/words/mumuni} \\
\hline
sharrap & pidgin & shut up & \url{http://naijalingo.com/words/sharrap} \\
\hline
tief & pidgin & thief & \url{http://naijalingo.com/words/tief} \\
\hline
tiff & pidgin & thief & \url{http://naijalingo.com/words/tiff} \\
\hline
waka jam & pidgin & an insult/curse towards you and loved ones & \url{http://naijalingo.com/words/waka-jam} \\
\hline
agba iya|awon agba iya & yoruba & older person, who despite his age, is still useless & \url{https://www.legit.ng/1031944-8-insults-yoruba-mothers-use-will-reset-brain.html} \\
\hline
agbaya & yoruba & derogatory word against old people & \\
\hline
agbero|agberos|awon agbero & yoruba & used to describe manual laborers from lower economic classes; sometimes deployed on twitter for ad hominem attacks & \url{https://en.wiktionary.org/wiki/agbero} \\
\hline
akpamo & yoruba & fool & \url{http://naijalingo.com/words/akpamo} \\
\hline
apoda|awon apoda & yoruba & who is confused, lost direction & \url{https://www.legit.ng/1031944-8-insults-yoruba-mothers-use-will-reset-brain.html} \\
\hline
arindin|awon arindi & yoruba & acts like an idiot & \url{https://www.nairaland.com/3237758/she-called-him-arindin-sitting} \\
\hline
arro & yoruba & stupid person & \url{http://naijalingo.com/words/arro} \\
\hline
atutupoyoyo & yoruba & ugly being & \url{http://naijalingo.com/words/atutupoyoyo} \\
\hline
ayama & yoruba & disgusting & \url{http://naijalingo.com/words/ayama} \\
\hline
ayangba & yoruba & prostitute & \\
\hline
didirin|awon didirin & yoruba & stupid & \url{https://www.legit.ng/1031944-8-insults-yoruba-mothers-use-will-reset-brain.html} \\
\hline
eyankeyan & yoruba & synonym to lasan & \citet{farinde2020socio} \\
\hline
lasan & yoruba & ordinary; when combined with a community name, may mean that this group is inferior to Yorubas & \citet{farinde2020socio} \\
\hline
malu|awon malu & yoruba & cow & \\
\hline
obun|awon obun & yoruba & dirty & \url{https://www.legit.ng/1031944-8-insults-yoruba-mothers-use-will-reset-brain.html} \\
\hline
ode|awon ode & yoruba & stupid & \url{https://www.legit.ng/1031944-8-insults-yoruba-mothers-use-will-reset-brain.html} \\
\hline
odoyo & yoruba & very stupid person & \url{http://naijalingo.com/words/odoyo} \\
\hline
ole|awon ole & yoruba & thief & \citet{udanor2019combating} \\
\hline
olodo|olodos|awon olodo & yoruba & stupid & \url{https://www.legit.ng/1031944-8-insults-yoruba-mothers-use-will-reset-brain.html} \\
\hline
oloshi|awon oloshi & yoruba & unfortunate, who does rubbish a lot, criminal & \url{https://www.legit.ng/1031944-8-insults-yoruba-mothers-use-will-reset-brain.html} \\
\hline
omo ale|awon omo ale & yoruba & bastard & \citet{farinde2020socio} \\
\hline
oponu|awon aoponu & yoruba & idiot & \url{https://www.legit.ng/1031944-8-insults-yoruba-mothers-use-will-reset-brain.html} \\
\hline
ota|awon ota & yoruba & enemy & \url{http://naijalingo.com/words/ota} \\
\hline
owo & yoruba & fool & \\
\hline
suegbe & yoruba & idiot & \url{http://naijalingo.com/words/suegbe} \\
\hline
werey|awon weyre & yoruba & crazy, mad & \url{http://naijalingo.com/words/werey} \\
\hline
yeye|awon akin yeye & yoruba & useless & \citet{udanor2019combating} \\
\hline
jeri & pidgin & fool & \url{http://naijalingo.com/words/jeri} \\
\hline
shalam & pidgin & & \\
\hline
biafrat|biafraud & combined & targeting Biafra & \\
\hline
fulanimal & combined & targeting Fulanis & \\
\hline
yorubastard|yariba|yorobber & combined & targeting Yorubas & \\
\hline
baby factory|baby factories & combined & targeting Igbo & \\
\hline
niyamiri & combined & & \\
\hline
\end{tabular}}
\caption{Slurs used in the Nigerian context}
\label{tab:list_hate_words}

\end{table*}

\paragraph{Community names}
Second, we define a list of names of communities that are often targeted by hate speech in Nigeria, again relying on past work \cite{onanuga2023arewaagainstlgbtq} and local knowledge from Nigerian colleagues.
We build an initial list (see Table \ref{tab:list_community_names} for the full list of considered and retained community names) and we then restrict this initial list of community names to the names that are most frequently mentioned on Nigerian Twitter. 
This approach yields 12 communities, including 5 ethnic groups (Yoruba, Igbo, Hausa, Fulani, Herdsmen), 2 religious groups (Christians and Muslims), 3 regional groups (Northern, Southern, Eastern) and 2 groups on gender identity and sexual orientation (Women and LGBTQ+). As mentioned earlier, Herdsmen are not a ethnic group per se but this term refers exclusively to Fulani herdsmen in the Nigerian context, hence the categorization as an ethnic group. For each of these groups, we list the different denominations of each group as well as their plural form and combine it in regular expressions (see Table \ref{tab:list_community_regexes}). Finally, we also identify 5 words combining a community name with a derogatory word (e.g., ``fulanimal'') that we coin \textit{combined word} thereafter. Since some targets are very rare in the annotated data, we decide to bundle the 12 communities into 5 groups: North (Northern, Hausa, Fulani, Herdsmen), South (Southern, Yoruba), East (Igbo, Biafra), Women, Muslim and Other (Christian, LGBTQ+). 

\begin{table}[h!]
\centering
\begin{tabular}{|l|l|l|}
\hline
\textbf{Community word} & \textbf{Frequency} & \textbf{Retained} \\
\hline
christian & 1.88E-03 & yes \\
muslim & 2.10E-03 & yes \\
northern & 1.25E-03 & yes \\
southern & 5.30E-04 & yes \\
hausa & 7.12E-04 & yes \\
fulani & 8.81E-04 & yes \\
yoruba & 1.37E-03 & yes \\
igbo & 1.52E-03 & yes \\
women & 4.93E-03 & yes \\
biafra & 1.60E-03 & yes \\
arewa & 1.30E-03 & yes \\
LGBTQ+ & 1.12E-03 & yes \\
herdsmen & 7.49E-04 & yes \\
eastern & 2.09E-04 & yes \\ \hline
tiv & 3.98E-05 & no \\
kanuri/beriberi & 1.82E-05 & no \\
ibibio & 1.45E-05 & no \\
ijaw/izon & 6.02E-05 & no \\
buharist & 1.15E-04 & no \\
ipobite & 6.22E-08 & no \\
arne & 3.82E-06 & no \\
transgender & 3.83E-05 & no \\
middle belt & 3.45E-05 & no \\
jukun & 6.93E-06 & no \\
Niger Delta & 2.42E-04 & no \\
yorubawa & 4.07E-07 & no \\
berom & 4.84E-05 & no \\
\hline
\end{tabular}
\caption{List of considered community words and their frequency in the Twitter dataset. The frequency for each word corresponds to the number of tweets containing the word divided by the total number of tweets.}
\label{tab:list_community_names}
\end{table}
\begin{table*}[h!]
\centering
\begin{tabular}{|l|l|}
\hline
\textbf{Community} & \textbf{Regular expression} \\
\hline
christian & christian|christians \\
muslim & muslim|muslims|islam|islamic \\
northern & northern|northerner|northerners|arewa|almajiri \\
southern & southern|southerner|southerners \\
hausa & hausa|hausas \\
fulani & fulani|fulanis \\
yoruba & yoruba|yorubas \\
igbo & igbo|ibo|ibos|igbos \\
women & women|woman|girl|girls|female|females \\
lgbt & lgbt|lgbtq|lgbtq+|gay|gays|lesbian|lesbians|transgender|transgenders \\
herdsmen & herdsmen|herdsman \\
eastern & eastern|easterner|easterners|biafra \\
\hline
\end{tabular}
\caption{Community Regex Mapping}

\label{tab:list_community_regexes}

\end{table*}


\begin{table}[h!]
\centering
\small
\resizebox{0.45\textwidth}{!}{
\begin{tabular}{|p{5cm}|c|} 
\hline
\multicolumn{1}{|c|}{\textbf{Word category}}                            & \textbf{Number of words} \\ \hline
Community names                                                         & 12                       \\ \hline
English hate words                                                      & 26                       \\ \hline
Non-English hate words                                                  & 58                       \\ \hline
Combined words                                                          & 5                        \\ \hline
Total number of hate words  (in all languages)                          & 84                       \\ \hline
Total number of hate words, including combined words (in all languages) & 89                       \\ \hline
\end{tabular}}
\caption{Summary statistics on the number of words per category}
\label{tab:summary_stats_word_lists}
\end{table}

\subsubsection{Sampling and evaluation sets}
We draw two distinct random samples of 100 million tweets each, one for sampling and model training $D_s$ and the other one for evaluation $D_e$.

\subsubsection{Stratified sample}\label{app:stratified-sample}

As previously stated, the extreme imbalance in our classification task makes random sampling ineffective and prohibitively expensive. With the aim to build high-performing classifiers at a reasonable cost, we build and annotate a stratified sample of tweets from $D_s$. 
We use three different sampling strategies to build this stratified sample. First, for each possible combination of community name and hate word, we sample up to 4 tweets that both contain the respective hate word and match with the respective community regular expression. The subset of tweets containing both the hate word and the community regular expression may be smaller than 4 and we sample the full subset in that case. Second, for each combined word W, we randomly sample 50 tweets containing W. Some combined words occur at a very low frequency such that the sample size is sometimes smaller than 50. Finally, for each community, we draw 50 random tweets matching the community regular expression, in order to avoid having a classifier that associates the community name with hate speech.

This yields a stratified sample of 1,607 tweets annotated as either hateful, offensive or neutral.

\subsubsection{Active learning sample}

Each active learning iteration samples a total of 100 tweets. The type of active learning method we employ is called \textit{certainty sampling} and consists in sampling instances at the top of the score distribution in order to annotate false positives and maximize precision. Specifically, each iteration $i$ consists of: 
\begin{itemize}
    \item Model training: we train a model on all of the labels we have, that is the stratified sample and the combination of all Active Learning samples from prior iterations
    \item Inference: we then deploy this model on $D_s$ and rank all tweets based on their BERT confidence score.
    \item Sampling and annotating: we define 5 rank buckets as: $[1,10^3], [10^3, 10^4], [10^4, 10^5], \newline
    [10^5, 10^6], [10^6, 10^7]$. We then sample $n$ tweets per rank bucket and annotate this sample.
    
\end{itemize}

We conduct a total of 25 iterations, of which 10 are conducted on the subset of $D_s$ containing community keywords and 15 on the full $D_s$. In our active learning process, three separate phases can be distinguished:

\begin{itemize}
    \item iterations 1-10: 
    \begin{itemize}
        \item the sampling is done on the subset of $D_s$ containing community words
        \item the active learning process is done separately for the hateful and the offensive classes
        \item the value of $n$ equals 10
        \item the overall sample size per iteration is 100 and equals to 5 buckets x n=10 x 2 classes (hateful and offensive)
    \end{itemize}
    \item iterations 10-19
    \begin{itemize}
        \item the sampling is done on the full sampling set $D_s$
        \item the active learning process is done separately for the hateful and the offensive classes
        \item the value of $n$ equals 10
        \item the overall sample size per iteration is 100 and equals to 5 buckets x n=10 x 2 classes (hateful and offensive)
    \end{itemize}
    \item iterations 20-24
    \begin{itemize}
        \item the sampling is done on the full sampling set $D_s$
        \item the active learning process is done only for the hateful class
        \item the value of $n$ equals 20
        \item the overall sample size per iteration is 100 and equals to 5 buckets x n=20 x 1 class (hateful)
    \end{itemize}    
\end{itemize}

\subsection{Annotation}\label{appendix:annotation}

\subsubsection{Annotation team}

The annotation team was composed of a Hausa man, a Hausa-Fulani woman, an Igbo man and a Yoruba woman.

\subsubsection{Annotation guidelines}
\label{app:annotation-guidelines}

\paragraph{Offensive tweets}
For tweets to be offensive, but not hateful, a tweet must satisfy all of the following criteria. 
\begin{itemize}
    \item The hate keyword is being used as pejorative towards another individual or group, and this group is not one of our communities. 
    \begin{itemize}
        \item  A personal attack against another individual, that does not mention a protected attribute such as, race, ethnicity, national origin, disability, religious affiliation, caste, sexual orientation, sex, gender identity and serious disease. 
        \item An insult towards a group based on non-protected attributes, such as, hobbies, fandom (e.g., sports, comic books).  
    \end{itemize}
    \item It is not offensive if the hate keyword is not being used on an individual or group.
    \begin{itemize}
        \item Not offensive if directed towards inanimate objects, abstract concepts (that do not have religious or cultural significance) or animals (unless the animal is used as a negative metaphor to describe a community). We define these as “out-of-scope entities” \cite{rottger-etal-2021-hatecheck}. 
    \end{itemize}
    \item It is not offensive if the hate word is self-referential. This would account for some types of sarcasm, or humour via self deprecation. 
    \item It is not offensive if the hate word is used for emphasis without being directed towards an individual or group. Several offensive words such as “shit” or “stupid” can be used as exclamations. 
    \item If the hate word is being used ambiguously (not recognizable as pejorative), then it is offensive if your answer is yes to one of these questions. 
    \begin{itemize}
        \item Can you imagine that someone might be offended by this? (err on the side of caution, aim for the lower bound)
        \item Would Twitter potentially detect it as an insult and make the user verify before posting?
    \end{itemize}
\end{itemize}

\paragraph{Hateful tweets}
This section is adapted from \cite{waseem-hovy-2016-hateful, basile-etal-2019-semeval} and Facebook Community Standards\footnote{https://transparency.fb.com/en-gb/policies/community-standards/hate-speech/}. For tweets to be hateful, instead of merely offensive, the tweet must satisfy one or more of the following criteria:

\begin{itemize}
    \item Uses a sexist, racial or homophobic slur.
    \begin{itemize}
        \item Misogyny/Sexist slurs to be defined as a statement that expresses hate towards women in particular (in the form of insulting, sexual harassment, threats of violence, stereotype, objectification and negation of male responsibility).
        \item Racial slurs to be defined as an insult that is designed to denigrate others based on their race or ethnicity. 
        \item Homophobic slurs to be defined as an insult that is designed to denigrate other on the basis of sexuality. This includes slurs targeted towards specific LGBTQ+ communities, such as transphobic slurs. 
        \item Usage of slur must not constitute a “reclaiming” of negative terms by the community in question. For instance, the n* word or “fag” or “bitch”. 
    \end{itemize}
    \item Attacks a minority.
    \begin{itemize}
        \item Minorities to be defined as a group based on protected characteristics: race, ethnicity, national origin, disability, religious affiliation, caste, sexual orientation, sex, gender identity and serious disease. 
        \item Attack to be defined as violent or dehumanizing speech, harmful stereotypes, statements of inferiority, expressions of contempt, disgust or dismissal, cursing and calls for exclusion or segregation.
        \item Seeks to silence a minority.
        \item Criticizes a minority (without a well founded argument).
        \begin{itemize}
            \item Criticizes a minority and uses a straw man argument.
            \item Blatantly misrepresents truth or seeks to distort views on a minority with unfounded claims.
        \end{itemize}
        \item Negatively stereotypes a minority. Negative stereotypes to be defined as dehumanizing comparisons that have historically been used to attack, intimidate, or exclude specific groups.
        \item Promotes, but does not directly use, hate speech or violent crime.
        \begin{itemize}
            \item Shows support of problematic hashtags. e.g.,“\#BanIslam”
            \item Defends xenophobia, racism, sexism, homophobia or other types of intolerance and bigotry.
        \end{itemize}
        \item If it is a retweet, it must indicate support for the original tweet. People sometimes share content that includes someone else’s hate speech to condemn it or raise awareness.
    \end{itemize}
\end{itemize}

\subsection{Language distribution}

\begin{table*}[]
\centering
\resizebox{0.6\textwidth}{!}{%
\begin{tabular}{|l|l|l|}
\hline
                                  & Stratified + active learning sets & Random set \\ \hline
English &   74.2 &  77   \\ \hline
English \& Nigerian Pidgin &   11 & 1.5   \\ \hline
English \& Yoruba &   4.2 &   -              \\ \hline
Nigerian Pidgin  &   3.6 &  7.3               \\ \hline
English \& Hausa  &  2.2 &  -               \\ \hline
Hausa  &  1 & 1.2                 \\ \hline
Yoruba  &  - & 1                 \\ \hline
URLs &  - & 6                 \\ \hline
Emojis  &  - & 2.3                 \\ \hline

\end{tabular}%
}
\caption{Share of each language across datasets (in \%). Hyphens indicate that the value is under 1\%.}
\label{tab:language_distribution}
\end{table*}
We ask the annotators to characterize the language of a random sample of 500 tweets, both for the stratified and active learning sets and for the random sample. We report the language distribution in Table \ref{tab:language_distribution}.

\subsection{Models}

\subsubsection{Number of parameters}
Conversational BERT has 110 million parameters. The XLM models, BERTweet and AfriBERTa have 125 million parameters. The DeBERTaV3 models have 86 million parameters. The number of parameters for GPT3.5 is undisclosed by OpenAI.

\subsubsection{Pretraining of NaijaXLM-T}
We followed \citet{alabi-etal-2022-adapting} and performed an adaptive fine tuning of XLM-R \cite{conneau-etal-2020-unsupervised} on our Twitter dataset. 
We kept the same vocabulary as XLM-R and trained the model for one epoch, using 1\% of the dataset as validation set. The training procedure was conducted in a distributed environment, for approximately 10 days, using 4 nodes with 4 RTX 8000 GPUs each, with a total batch size of 576.

\subsubsection{Supervised Learning}\label{app:supervised_learning}

\paragraph{Hyperparameter tuning}

Hyperparameter tuning was conducted in a 5-fold cross validation training. A grid search was run testing different learning rates (from 1e-5 to 5e-5). The cross validation trainings were conducted for 10 epochs. The batch size used was 8, and three different seeds were used for each learning rate. We used F1-score as early stopping metric for hate speech detection models. The best results were averaged across the seeds, and the best combination after the grid search was picked as the resulting model.

\paragraph{Computing infrastructure}

For supervised learning, we used either V100 (32GB) or RTX8000 (48GB) GPUs for finetuning. The average runtime for finetuning is 45 minutes. Inferences from off-the-shelf models were ran locally on a laptop CPU.

\subsubsection{Off-the-shelf models}\label{app:off_the_shelf_models}

\paragraph{Perspective API}

We used the IDENTITY\_ATTACK category for HSD with Perspective API as it is the closest to our hate speech definition. This is a binary classification problem and the API outputs a score between 0 and 1. The inferences were run on February 1, 2024.

\paragraph{GPT3.5}\label{llms_prompts}

We use the \textit{gpt-3.5-turbo-0613} model. The prompt used for zero-shot predictions with this model is: \textit{"Now consider this message : '[TWEET]' Respond 0 if this message is neutral, 1 if this message is offensive and 2 if this message is hateful. It is very important that you only respond the number (e.g., '0', '1' or '2')."}

The prompt is run 5 times for each tweet. We then define the hateful score as the share of the 5 times for which the model predicted that the tweet was hateful. We then use this score to compute the average precision. 
We use all default values for the main hyperparameters, including 1 for temperature.

\subsubsection{Evaluation results}

\begin{figure}[h!]
    \centering
        \includegraphics[width=7.75cm]{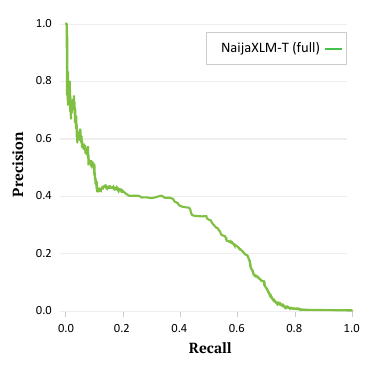}
\caption{Precision-recall curve on the random set}
\label{fig:pr_curve_random}
\end{figure}

\begin{figure*}[h!]
    \centering
        \includegraphics[width=16cm]{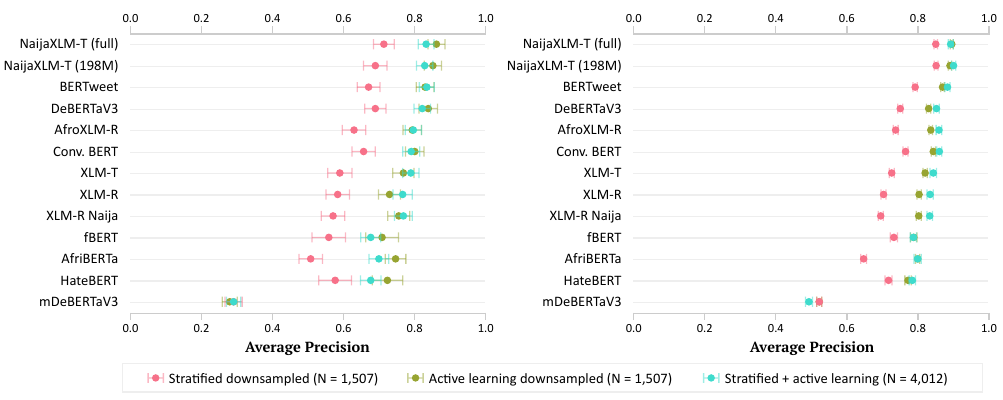}
\caption{Average precision on the holdout and top-scored sets across models trained on the downsampled stratified set, the downsampled active learning set and the full training set, composed of the stratified and active learning sets. Error bars indicate 95\% bootstrapped confidence intervals.}
\label{fig:diversity_rest}
\end{figure*}

We provide the diversity results for the holdout and the top-scored sets in Figure \ref{fig:diversity_rest}. We also provide the precision-recall curve for NaijaXLM-T on the random sample in Figure \ref{fig:pr_curve_random}.


\subsubsection{Human-in-the-loop moderation}\label{app:human-in-the-loop}

In the human-in-the-loop setting, we compute the proportion of all tweets to be reviewed by human annotators as follows. We first compute the number of predicted positives PP:
$$PP = TP + FP = TP/\textrm{precision}$$
$$PP = (\textrm{recall} * (TP+FN))/\textrm{precision}$$
$$PP = (\textrm{recall} * \textrm{total \# hateful tweets})/\textrm{precision}$$
$$PP = (\textrm{recall} * \textrm{base rate} * \textrm{total \# tweets})/\textrm{precision}$$

We then derive the share $S$ of all tweets that are predicted positive by a given model, that is the share of all tweets that will be reviewed by human moderators in a human-in-the-loop approach, by dividing $PP$ by the total number of tweets:

$$S = (\textrm{recall} * \textrm{base rate})/\textrm{precision}$$

with the base rate equal to the prevalence of hateful content which is 0.16\%. We finally use the precision and recall values from the precision-recall curve to derive the curve illustrating the relationship between recall and $S$. 

\end{document}